\documentclass[10pt,twocolumn,letterpaper]{article}

\usepackage{cvpr}              

\usepackage{kotex}
\usepackage{multirow}

\usepackage{xcolor}
\definecolor{deepblue}{RGB}{0,70,160}

\definecolor{cvprblue}{rgb}{0.21,0.49,0.74}
\usepackage[pagebackref,breaklinks,colorlinks,allcolors=cvprblue]{hyperref}
\usepackage{graphicx}
\def\paperID{6837} 
\def\confName{CVPR}
\def\confYear{2026}

\title{BindEdit: Taming Attention Leakage for Precise Multi-Object Image Editing}

\author{
Chaewon Park$^{1}$ \enspace
Soyoon Lee$^{1}$ \enspace
Naeun Lee$^{2}$ \enspace
Minjung Shin$^{2^{*}}$ \enspace
Seogkyu Jeon$^{3^{*}}$ \enspace
Kibeom Hong$^{1^{*}}$\\
$^{1}$Sookmyung Women's University \quad
$^{2}$Yonsei University \quad
$^{3}$Samsung Research\\
{\tt\footnotesize
\{sea9392, synuo, kb.hong\}@sookmyung.ac.kr \enspace
\{skdms1449, smj139052\}@yonsei.ac.kr \enspace
jone9312@gmail.com
}
}

\begin{document}
\maketitle
\renewcommand{\thefootnote}{\fnsymbol{footnote}}
\footnotetext[1]{denotes the corresponding author.}
\renewcommand{\thefootnote}{\arabic{footnote}}
\begin{abstract}
Real image editing enables precise manipulation of visual content, yet existing methods often fail in complex multi-object scenarios, causing semantic blending, object duplication, or incomplete edits. We attribute these failures to attention leakage, where signals across spatial regions and text tokens become entangled during the denoising process. Specifically, we identify two distinct forms of leakage:  Edit-Token Leakage, where ambiguous token-region alignment leads to object blending, and Source Dominance Leakage, where tokens of unchanged source objects overwhelm the attention intended for target entities. To resolve these leakages, we propose \textbf{BindEdit}, which enforces attention-level constraints within a single diffusion trajectory. To suppress Edit-Token Leakage, BindEdit jointly regularizes cross- and self-attention so that each target token group is bound to its corresponding spatial region while maintaining instance-level separation. To suppress Source Dominance Leakage, a cross-attention re-balancing mechanism amplifies target token influence and attenuates residual source semantics within editable regions. Moreover, a region fidelity term ensures that each target concept is expressed coherently across the entire editing mask. Additionally, we propose a comprehensive multi-object benchmark encompassing diverse object counts and categories. Extensive experiments demonstrate that BindEdit consistently outperforms existing methods within a single diffusion trajectory, maintaining robust performance across both single- and multi-object editing scenarios.
\end{abstract}

\section{Introduction}
\label{sec:intro}

\begin{figure}[t]
\centering
  \includegraphics[width=\linewidth, page=1]{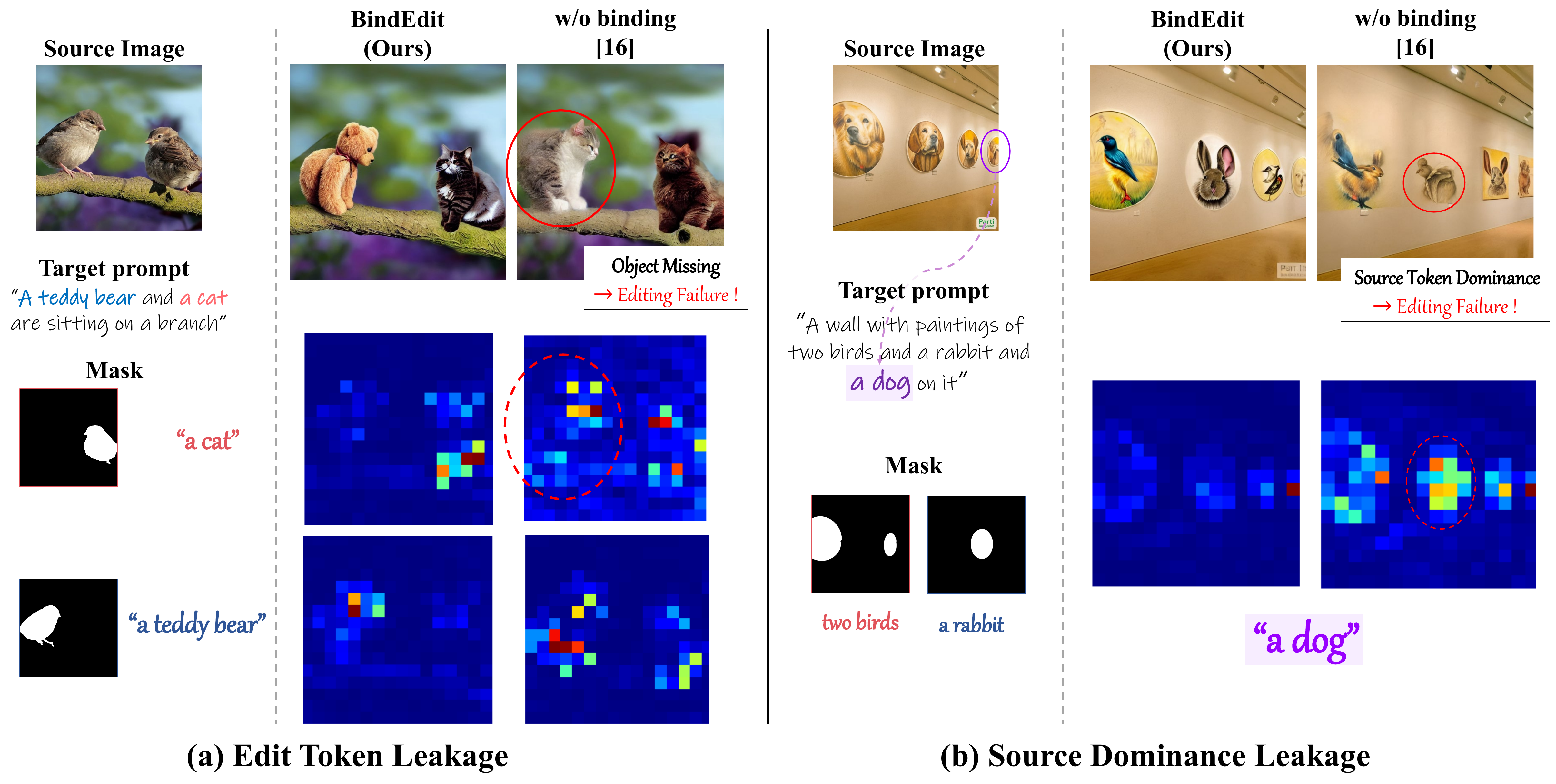}
  \caption{(a) \textbf{Edit-Token Leakage}: when editing two birds into a teddy bear and a cat, the cross-attention maps of target tokens both activate across unintended regions without our binding guidance, causing one object to go missing. BindEdit precisely binds each target token group to its designated mask region. (b) \textbf{Source Dominance Leakage}: when editing three of four dog paintings into two birds and a rabbit, the source token ``a dog'' maintains strong attention bias that spill over into the editable areas (red circle), suppressing target token signals and retaining canine features. BindEdit suppresses residual source influence, enabling faithful semantic replacement.}
  \label{fig:teaser}
\end{figure}
Real image editing is essential in many industries as it enables direct modification of existing visual content. With the advent of diffusion models~\cite{ho2020denoising,song2020denoising,Rombach2021HighResolutionIS,Saharia2022PhotorealisticTD,Peebles2022ScalableDM,Podell2023SDXLIL,Labs2025FLUX1KF}, editing can be performed in a fine-grained manner while preserving the realism and structure of unedited regions. Among these approaches, text-guided editing is particularly compelling due to its intuitive natural language interface. Specifically, text-guided image editing modifies regions relevant to a target prompt within a source image, simultaneously preserving the integrity of the remaining regions.

While previous text-guided editing methods~\cite{hertz2022prompt,tumanyan2023plug,Liu2024TowardsUC,mokady2023null,huberman2024edit,Wu2024FreeDiffPF,Koo2024FlexiEditFL,hertz2023delta,Nam2023ContrastiveDS,ren2025fds,Kulikov2024FlowEditIT,Avrahami2024StableFV,Shin2025ExploringMD} primarily focus on single-object scenarios, real-world images commonly contain multiple distinct entities requiring simultaneous editing. This necessity has led to the emergence of \emph{multi-object image editing}, defined as simultaneously and precisely modifying multiple independent entities. However, existing single-object approaches often fail in these complex settings. For instance, as illustrated in Fig.~\ref{fig:teaser}, when the source image contains two birds to be edited into a cat and a teddy bear respectively, the resulting objects exhibit blended identities rather than distinct transformations. Similarly, when the target prompt retains source-related tokens such as ``a dog'', the edited regions often preserve the appearance of the original entities instead of reflecting the intended target concepts. These failures highlight a fundamental gap between single-object and multi-object editing that current methods struggle to bridge.

We attribute these persistent limitations to \emph{attention leakage}, a phenomenon widely recognized in text-conditioned generation~\cite{Li2023DivideB,Dahary2024BeYB,Han2025SpatialTO,Chen2023TrainingFreeLC,Park2025CrossAttentionHP} that remains underexplored in the context of multi-object editing. Unlike single-object scenarios, multi-object image editing necessitates that each target prompt be precisely coupled with its corresponding source entity to ensure faithful modification. However, in standard architectures, cross-attention is jointly calculated across all spatial locations and text tokens, thereby conflating object-specific signals and hindering independent control, even when per-object masks and prompts are provided. This entangled attention prevents the model from isolating edits to their intended targets, ultimately leading to the aforementioned failure cases. To address this, previous multi-object editing methods~\cite{chakrabarty2024lomoe,yang2023object,Fu2025LayerEditDM, Huang2024ParallelEditsEM} have attempted to circumvent the leakage by decomposing the task into a fragmented generation process. These approaches perform iterative single-object operations followed by a merging step. While intuitive, this multi-pass denoising strategy introduces significant limitations in terms of both fidelity and efficiency. By isolating each object into disjointed generation steps, these methods often result in overly constrained to per-object masks. As a result, edited regions appear as unnaturally stitched patches with visible boundary artifacts, rather than being harmoniously synthesized with surrounding content. Furthermore, the sequential nature of such pipelines incurs prohibitive computational overhead, which limits their scalability and practical utility for complex images.

Motivated by these observations, we conduct a rigorous analysis of \textit{how attention leakage manifests during the transition from the inverted source image to the target prompt}, and identify two distinct forms of leakage (Fig.~\ref{fig:teaser}).
The first, \textit{Edit-Token Leakage}, refers to the entanglement of attention signals across different target objects that causes their identities to blend rather than being edited independently. As illustrated in Fig.~\ref{fig:teaser}.(a), an attempt to modify ``two birds'' into ``a teddy bear and a cat'' yields a single blended object combining features of both. This arises because target tokens alone cannot specify which spatial regions in the source image they should attend to, even when only object-related words change from the source prompt. The second, \textit{Source Dominance Leakage}, occurs when the source image contains multiple objects of the same category and only a subset is targeted for editing. As shown in Fig.~\ref{fig:teaser}.(b), among four dogs in the source image only three are intended to be edited into two birds and a rabbit, yet the edited regions still exhibit canine features. This arises because source tokens describing unedited objects maintain strong attention bindings to their original spatial regions and these bindings spill over into the editable areas, suppressing the target token signals.

To this end, we propose \textbf{BindEdit}, a framework designed to precisely bind each target prompt to its corresponding spatial region through attention-level constraints within a single diffusion trajectory. BindEdit supervises both cross-attention and self-attention using region masks, enabling each object to be edited independently while preserving the coherence of the overall image. To resolve \emph{Edit-Token Leakage}, we introduce a joint regularization on cross- and self-attention. In cross-attention, each target token group is guided to concentrate its activation within its designated semantic region. In self-attention, spatial tokens within each target instance are encouraged to primarily interact with one another, reinforcing instance-level separation. By jointly enforcing both semantic-level binding and instance-level separation, this regularization ensures that each object is edited without interfering with others. BindEdit further tackles \emph{Source Dominance Leakage} by suppressing source and background token activations that persist from the inversion process within editable regions, allowing target tokens to dominate the cross-attention distribution. This ensures that the edited regions are driven by the intended target semantics rather than inheriting residual source characteristics. Additionally, we enforce spatially consistent attention distribution within each editing mask, ensuring that each object is expressed coherently across its entire region and avoiding fragmented generations.

In addition, we introduce a comprehensive benchmark to facilitate standardized evaluation in complex multi-object scenarios. While existing benchmarks~\cite{chakrabarty2024lomoe,yang2023object} mainly address editing with fewer objects, our benchmark extends to three or more objects and covers multiple object categories and diverse spatial layouts. Extensive experiments demonstrate that BindEdit consistently outperforms existing methods within a single diffusion trajectory, maintaining robust performance across both single- and multi-object editing scenarios.
\section{Related works}
\label{sec:rel}
\subsection{Diffusion-based Text-to-Image Editing}
Diffusion-based image editing~\cite{hertz2022prompt, tumanyan2023plug, mokady2023null,huberman2024edit,Meng2021SDEditGI,Couairon2022DiffEditDS,Cao2023MasaCtrlTM,Liu2024TowardsUC,Wu2024FreeDiffPF,Koo2024FlexiEditFL, hertz2023delta, ren2025fds} has been actively studied following the success of text-to-image generation models~\cite{ho2020denoising, song2020denoising,Rombach2021HighResolutionIS, Saharia2022PhotorealisticTD, Peebles2022ScalableDM, Podell2023SDXLIL,Labs2025FLUX1KF}. Early works rely on mask-based or region-constrained editing, such as SDEdit~\cite{Meng2021SDEditGI} and DiffEdit~\cite{Couairon2022DiffEditDS}, while subsequent approaches manipulate cross-attention maps to align generation with the target prompt, as in Prompt-to-Prompt~\cite{hertz2022prompt} and Plug-and-Play~\cite{tumanyan2023plug}. However, these methods implicitly assume a single editing target and lack mechanisms to separate multiple instances of the same class or coordinate region-specific edits, limiting their applicability to multi-object editing.
\subsection{Multi-object image editing}
Several approaches~\cite{yang2023object,chakrabarty2024lomoe,Brack2023LEDITSLI,Huang2024ParallelEditsEM,Li2025MoEditOL,Kang2025BoundedEM,Zhu2025MDEEditMD,Yoon2025SplitFlowFD,Fu2025LayerEditDM} extend diffusion-based image editing to multi-object scenarios. Early attempts decompose multi-object editing into independent single-object edits: LoMOE~\cite{chakrabarty2024lomoe}, OIR~\cite{yang2023object}, SplitFlow~\cite{Yoon2025SplitFlowFD}, and LayerEdit~\cite{Fu2025LayerEditDM} follow a separate-and-merge paradigm that isolates each object during editing and combines the results afterward. While reducing direct interference, these methods require object-wise noise prediction or multiple denoising runs, incurring computational cost that scales with object count and often degrading global coherence. To reduce this overhead, LEDITS++~\cite{Brack2023LEDITSLI} and ParallelEdits~\cite{Huang2024ParallelEditsEM} aggregate object-specific guidance signals within a shared diffusion trajectory, but still rely on object-wise computation without explicitly regulating attention interactions, leaving attention leakage unaddressed. In contrast, by enforcing attention-level constraints that bind each target prompt to its corresponding spatial region, BindEdit achieves coherent edits without object-wise noise separation.
\subsection{Attention Control in Diffusion Models}
Recent studies in multi-object text-to-image generation~\cite{Chefer2023AttendandExciteAS,Chen2023TrainingFreeLC,Dahary2024BeYB,Qiu2024SelfCrossDG,Chen2024RegionAwareTG,Han2025SpatialTO} have shown that attention distributions can be explicitly optimized to improve object localization and semantic grounding, for example by applying object-specific masks to suppress semantic leakage across subjects~\cite{Dahary2024BeYB} or aligning attention maps to target layouts via optimal transport~\cite{Han2025SpatialTO}. However, in image editing, attention is persistently biased by the source image. This introduces an additional challenge absent in generation: misalignment stems not only from interference among target tokens but also from residual source dominance, which these generation-oriented methods do not address. BindEdit bridges this gap by adapting attention-level guidance to the editing setting, jointly suppressing both forms of leakage within a single denoising process.

\section{Proposed methods} 
\label{sec:method}
\begin{figure*}[t]
\centering
  \includegraphics[width=\linewidth]{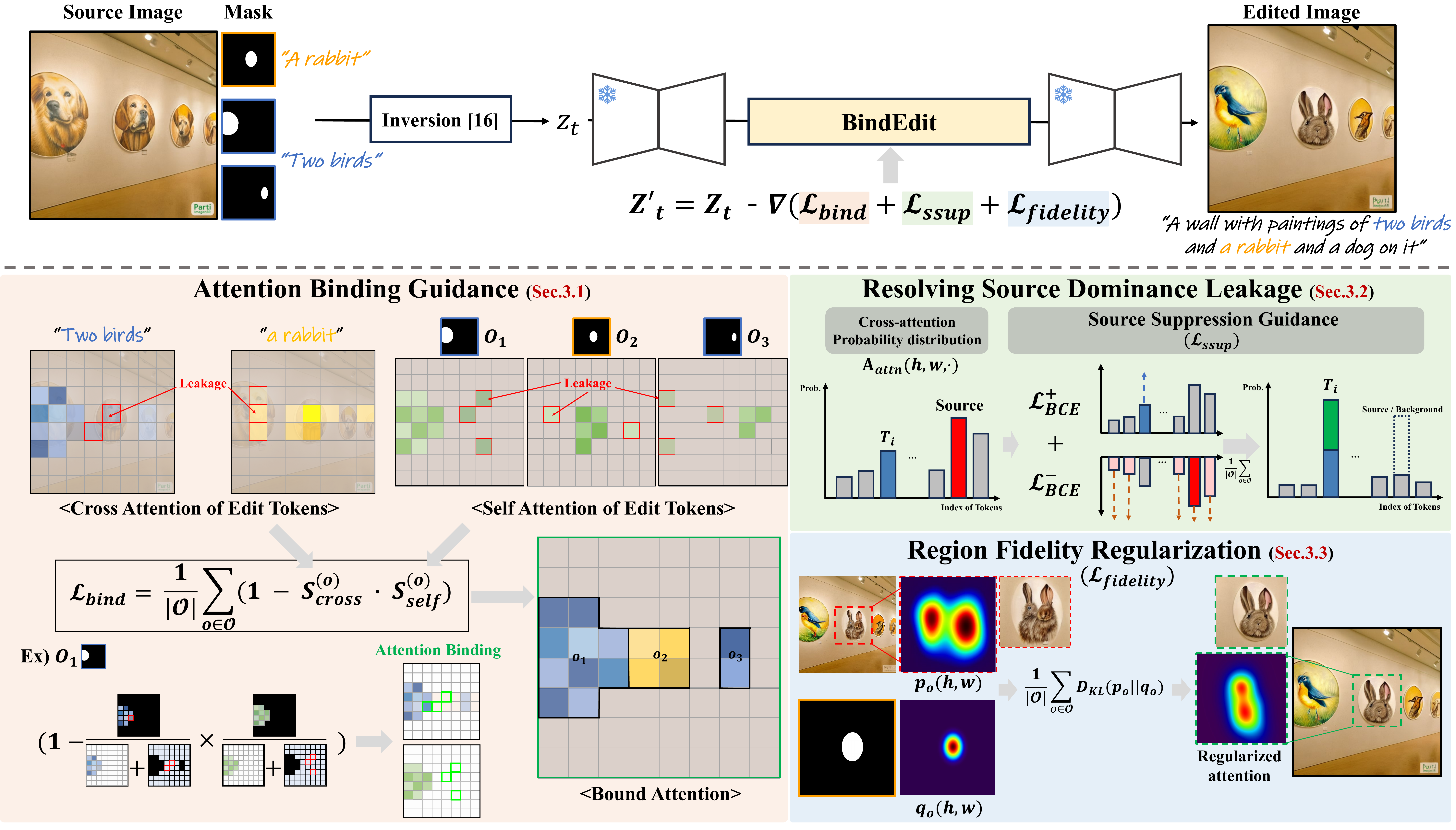}
  \caption{
    \textbf{Overview of the proposed BindEdit.} BindEdit guides the denoising process through three attention-level losses within a single diffusion trajectory. First, Attention Binding Guidance ($\mathcal{L}_{\text{bind}}$) jointly constrains cross- and self-attention to enforce semantic-level binding and instance-level separation. Second,  Source Suppression Guidance ($\mathcal{L}_{\text{supp}}$) re-balances cross-attention within editable regions by strengthening target tokens and mitigating source dominance. Lastly, Region Fidelity ($\mathcal{L}_{\text{fidelity}}$) consolidates scattered attention peaks into a single coherent mode via a truncated Gaussian prior.}
  \label{fig:overview}
\end{figure*}

Our goal is to enable simultaneous editing of multiple objects within a single diffusion trajectory while preventing attention leakage across objects. We present \textbf{BindEdit}, a training-free framework that achieves this by directly steering the intermediate latent variables $z_t$ throughout the denoising process without modifying the parameters of the pretrained denoising network. We adopt the edit-friendly DDPM inversion~\cite{huberman2024edit} to obtain the initial noisy latent for editing. As shown in Fig.~\ref{fig:overview}, BindEdit addresses the two identified forms of attention leakage through dedicated guidance losses: \textbf{Attention Binding Guidance} ($\mathcal{L}_{\text{bind}}$) for Edit-Token Leakage and \textbf{Source Dominance Suppression} ($\mathcal{L}_{\text{supp}}$) for Source Dominance Leakage. We further introduce a \textbf{Region Fidelity} term ($\mathcal{L}_{\text{\textbf{fidelity}}}$) to ensure spatially coherent generation within each editing mask.

Throughout this section, we denote the cross-attention score tensor as $\mathbf{A}_{\text{attn}} \in \mathbb{R}^{(H \times W) \times N_t}$, where each row is a probability distribution over $N_t$ text tokens at spatial location $(h,w)$. For a target token group $T_o$, we define its aggregated spatial attention map as $A_{T_o}(h,w) = \sum_{t \in T_o} \mathbf{A}_{\text{attn}}(h,w,t)$. 


\subsection{Attention Binding Guidance: Suppressing Edit-Token Leakage}
Simultaneous multi-object editing requires that each target token group binds precisely to its corresponding mask region at the attention level. When this binding fails, target tokens activate beyond their intended regions in both cross- and self-attention, producing blended or hybrid objects. As depicted in the bottom-left of Fig.~\ref{fig:overview}, the attention weights for ``two birds'' erroneously spread not only within the designated area but also into background regions and the mask for ``a rabbit,'' disrupting the correspondence between target concepts and their spatial regions.
This Edit-Token Leakage stems from both attention mechanisms. In cross-attention, target token groups fail to concentrate on their designated semantic regions, causing editing signals to be delivered to incorrect objects. In self-attention, spatial tokens from different instances freely interact with one another, breaking instance-level separation and inducing identity blending even across objects of different categories.

To address both forms of leakage simultaneously, we introduce \textbf{Attention Binding Guidance} ($\mathcal{L}_{\text{bind}}$), which enforces semantic-level binding through cross-attention and instance-level isolation through self-attention. Inspired by~\cite{Dahary2024BeYB}, we formulate a ratio-based objective that measures how strongly attention concentrates within the target region. For each object $o$, we define its concentration ratio $S^{(o)}$ as:
\begin{equation}
S^{(o)} = \frac{\mathcal{W}^{(o)}_{\text{in}}}{\mathcal{W}^{(o)}_{\text{in}} + \mathcal{W}^{(o)}_{\text{out}} + \varepsilon},
\end{equation}
where $\mathcal{W}^{(o)}_{\text{in}}$ and $\mathcal{W}^{(o)}_{\text{out}}$ denote the total sum of attention weights inside and outside the target region respectively. As $S^{(o)}$ approaches one, the out-of-region attention is driven toward zero. Since cross- and self-attention exhibit leakage in different ways, we define $\mathcal{W}^{(o)}_{\text{in}}$ and $\mathcal{W}^{(o)}_{\text{out}}$ independently for each mechanism.

\noindent\textbf{Cross-attention binding.}
For the cross-attention concentration ratio $S^{(o)}_{\text{cross}}$, we compute $\mathcal{W}^{(o)}_{\text{in}}$ as the sum of attention weights of the target token group within the object mask $\mathcal{M}_{\text{obj}}^{(o)}$, and $\mathcal{W}^{(o)}_{\text{out}}$ as the sum in the region outside the corresponding semantic mask $(1 - \mathcal{M}_{\text{sem}}^{(c)})$. Here $\mathcal{M}_{\text{sem}}^{(c)}$ denotes the union of all instance masks belonging to the same semantic category $c$. This formulation penalizes attention that leaks into semantically unrelated regions while permitting activation within the same category, allowing the model to maintain natural inter-instance context. Maximizing $S^{(o)}_{\text{cross}}$ ensures that each target token group is semantically bound to its designated region.

\noindent\textbf{Self-attention binding.}
For the self-attention concentration ratio $S^{(o)}_{\text{self}}$, we compute $\mathcal{W}^{(o)}_{\text{in}}$ by accumulating attention scores among spatial tokens within the target object region, and $\mathcal{W}^{(o)}_{\text{out}}$ by accumulating scores between the target region and non-target regions. Maximizing $S^{(o)}_{\text{self}}$ enforces instance-level isolation, ensuring that spatial features within each object primarily interact with one another rather than blending across instances.

\noindent\textbf{Joint formulation.}
Semantic binding and instance isolation must hold simultaneously: correct token-to-region mapping is ineffective if spatial features freely mix across instances, and vice versa. A simple summation would allow one term to compensate for the other. We therefore combine the two ratios as a product and average over all edited objects:
\begin{equation}
\mathcal{L}_{\text{bind}} = \frac{1}{|\mathcal{O}|}\sum_{o \in \mathcal{O}} \left(1 - S^{(o)}_{\text{cross}} \cdot S^{(o)}_{\text{self}}\right),
\end{equation}
where $\mathcal{O}$ is the set of all objects to be edited. This coupling ensures that neither condition alone is sufficient: the loss decreases only when both semantic binding and instance isolation are jointly satisfied.

All attention statistics are computed at the $16 \times 16$ spatial resolution, where object-level semantics are most prominently formed~\cite{Tang2022WhatTD,Tian2023DiffuseAA,Chefer2023AttendandExciteAS,Dahary2024BeYB,Han2025SpatialTO}. Since attention naturally concentrates within the intended regions as denoising progresses, persistent regularization at later stages risks over-constraining the synthesis and reducing image fidelity. To address this, we introduce a timestep-dependent scaling factor:
\begin{equation}
\lambda_t = 1 - \frac{\sum_{h,w}\mathcal{M}_{\text{obj}}^{(o)}(h,w) \cdot A_{T_o}(h,w)}{\sum_{h,w} A_{T_o}(h,w) + \varepsilon}, \quad \mathcal{L}_{\text{bind}}^{(t)} = \lambda_t \, \mathcal{L}_{\text{bind}}.
\end{equation}
When attention is diffused in early timesteps, $\lambda_t$ is close to one and the guidance provides strong binding. As more attention mass settles into the correct region, $\lambda_t$ decreases and the regularizer gracefully recedes. We provide a detailed derivation showing that $\lambda_t$ acts as a per-object gradient scaling coefficient in the supplementary material.

\subsection{Source Suppression Guidance: Resolving Source Dominance Leakage}
\label{sec:supp}
As identified in Sec.~\ref{sec:intro}, Source Dominance Leakage arises when multiple objects of the same category appear in the source image and only a subset is targeted for editing. This is particularly severe during the early denoising timesteps when source and target share similar semantics (\eg, a dog edited into a teddy bear), as source tokens dominate the cross-attention distribution and inhibit the intended target concepts from being expressed.
Unlike Edit-Token Leakage, which involves entanglement across target tokens, Source Dominance Leakage requires directly reshaping the cross-attention distribution within the editable region to shift probability mass from source tokens toward target tokens. Since cross-attention scores at each spatial location form a probability distribution over text tokens, binary cross-entropy provides a natural supervision signal that allows independent control over positive (target) and negative (source/background) terms.



We extract the attention within the target editing area using the object mask $\mathcal{M}_{\text{obj}} \in \{0,1\}^{H \times W}$ and define a binary token supervision mask $\mathcal{M}_{T_o} \in \{0,1\}^{N_t}$ that selects the target token group $T_o$ and sets all other entries to zero. We construct a contrastive binary cross-entropy loss:
\begin{equation}
\mathcal{L}_{\text{supp}} = \mathcal{L}_{\text{BCE}}^{+} + \gamma\,\mathcal{L}_{\text{BCE}}^{-}, \quad \gamma > 1,
\end{equation}
where the positive and negative terms are defined as
\begin{equation}
\begin{aligned}
\mathcal{L}_{\text{BCE}}^{+}
&= -\sum_{(h,w) \in \mathcal{M}_{\text{obj}}} \sum_{t \in T_o}
    \log\bigl(\mathbf{A}_{\text{attn}}(h,w,t) + \varepsilon\bigr), \\
\mathcal{L}_{\text{BCE}}^{-}
&= -\sum_{(h,w) \in \mathcal{M}_{\text{obj}}} \sum_{t \notin T_o}
    \log\bigl(1 - \mathbf{A}_{\text{attn}}(h,w,t) + \varepsilon\bigr).
\end{aligned}
\end{equation}

The positive term $\mathcal{L}_{\text{BCE}}^{+}$ increases the probability assigned to the target token group inside the editable region, while the negative term $\mathcal{L}_{\text{BCE}}^{-}$ penalizes attention allocated to all other tokens including source-related and background tokens. Since generic source tokens tend to dominate the initial distribution, we amplify the negative term by a factor $\gamma$ to effectively suppress residual source influence. The final loss averages over all edited objects:
\begin{equation}
\mathcal{L}_{\text{supp}} = \frac{1}{|\mathcal{O}|}\sum_{o \in \mathcal{O}} \mathcal{L}_{\text{supp}}^{(o)}.
\end{equation}
This supervision acts only on the cross-attention distributions within the editing masks and does not alter the global layout inherited from inversion, thereby driving semantic replacement in the editable regions while preserving the overall scene geometry. The robustness to parameter $\gamma$ is further analyzed in Sec.~\ref{sec:ablation}.

\subsection{Region Fidelity Regularization}
\label{sec:rf}
While the preceding two losses effectively suppress attention leakage across objects, the spatial distribution of target attention within each editing mask can be further improved with additional regularization. Specifically, we observe that target token activation can become sparse and fragmented within the editing mask in certain cases. Instead of forming a single coherent distribution, the activation splits into multiple scattered peaks. As shown in Fig.~\ref{fig:overview}, this can lead to multiple instances of the target concept appearing within a single mask region, such as two rabbits generated inside an area intended for one.

To further refine the editing quality in such cases, we introduce a lightweight region fidelity term. We normalize the attention inside the object mask $\mathcal{M}_{\text{obj}}^{(o)}$ as a spatial probability distribution:
\begin{equation}
p_o(h, w) = \frac{\mathcal{M}_{\text{obj}}^{(o)}(h, w)\, A_{T_o}(h, w)}{\sum_{h',w'} \mathcal{M}_{\text{obj}}^{(o)}(h', w')\, A_{T_o}(h', w') + \varepsilon}.
\end{equation}
We then construct a Gaussian reference $q_o(h, w)$ centered at the centroid of $\mathcal{M}_{\text{obj}}^{(o)}$ with variance covering the mask extent, truncated by the mask boundary and renormalized. Region fidelity is enforced via $D_{\text{KL}}(p_o \| q_o)$:
\begin{equation}
\mathcal{L}_{\text{fidelity}} = \frac{1}{|\mathcal{O}|}\sum_{o \in \mathcal{O}} \sum_{h,w} \mathcal{M}_{\text{obj}}^{(o)}(h, w)\, p_o(h, w) \log \frac{p_o(h, w)}{q_o(h, w) + \varepsilon}.
\end{equation}
This formulation penalizes $p_o$ whenever it places mass at isolated sub-regions, consolidating scattered peaks into a single coherent mode. To further sharpen this focus, we implement a thresholding mechanism that eliminates any non-global maxima, ensuring a definitive semantic assignment within the mask. While not essential for the majority of editing scenarios, this term provides a meaningful quality boost in difficult cases by ensuring that the target concept is expressed as a single complete object across the entire mask.

\subsection{Overall Objective}
\label{sec:objective}

The full editing objective combines the three attention-level constraints:
\begin{equation}
\mathcal{L}_{\text{edit}} = \lambda_{\text{bind}}\, \mathcal{L}_{\text{bind}}^{(t)} + \lambda_{\text{supp}}\, \mathcal{L}_{\text{supp}} + \lambda_{\text{fidelity}}\, \mathcal{L}_{\text{fidelity}},
\end{equation}
where $\lambda_{\text{bind}}$, $\lambda_{\text{supp}}$, and $\lambda_{\text{fidelity}}$ balance the three terms. During guided sampling, we directly optimize the latent state at selected timesteps without modifying the diffusion model. For a noisy latent $\mathbf{z}_t$, we apply a gradient step of the form:
\begin{equation}
\mathbf{z}_t \leftarrow \mathbf{z}_t - \eta\, \nabla_{\mathbf{z}_t} \mathcal{L}_{\text{edit}},
\end{equation}
where $\eta$ is the step size for latent optimization. This update steers the denoising trajectory toward attention patterns that respect object boundaries, suppress residual source semantics, and cover each editable region coherently, while the generative prior of the pretrained diffusion model remains fully intact.

\section{Experiments}
\label{sec:exp}
\subsection{Datasets and Implementation Details}
\label{sec:dataset}
\noindent\textbf{Datasets.}
We evaluate BindEdit on two established multi-object editing benchmarks and a newly proposed benchmark designed for more complex scenarios.
\textbf{LoMOE-Bench}~\cite{chakrabarty2024lomoe} contains 64 images with 2--7 annotated object masks and paired source--target prompts. While suitable for multi-object evaluation, most samples focus on minor attribute changes rather than large-scale compositional edits.
\textbf{OIR-Bench}~\cite{yang2023object} provides 100 multi-object text--image pairs that serve as a complementary testbed for assessing structural preservation under diverse object configurations.

\noindent\textbf{Extended Multi-Object Benchmark (Ours).}
To complement these benchmarks with more complex scenarios, we introduce an extended benchmark comprising 238 images that uniformly covers edit object counts ranging from 1 to 5 or more. This balanced distribution enables a controlled analysis of how each method scales with the number of simultaneously edited objects, providing a rigorous evaluation of semantic disentanglement and region-aware fidelity. Please refer to the supplementary material for more comprehensive details and examples of the proposed benchmark.

\noindent\textbf{Implementation details.}
BindEdit is a training-free method that requires no finetuning of the pretrained diffusion model. We set the loss balancing coefficients to $\lambda_{\text{bind}} = 20$, $\lambda_{\text{supp}} = 10$, and $\lambda_{\text{fidelity}} = 2$. All experiments are conducted on a single NVIDIA A6000 GPU. Further details including the latent optimization schedule are provided in the supplementary material.



\begin{figure*}[t!]
\centering
  \includegraphics[width=0.8\linewidth]{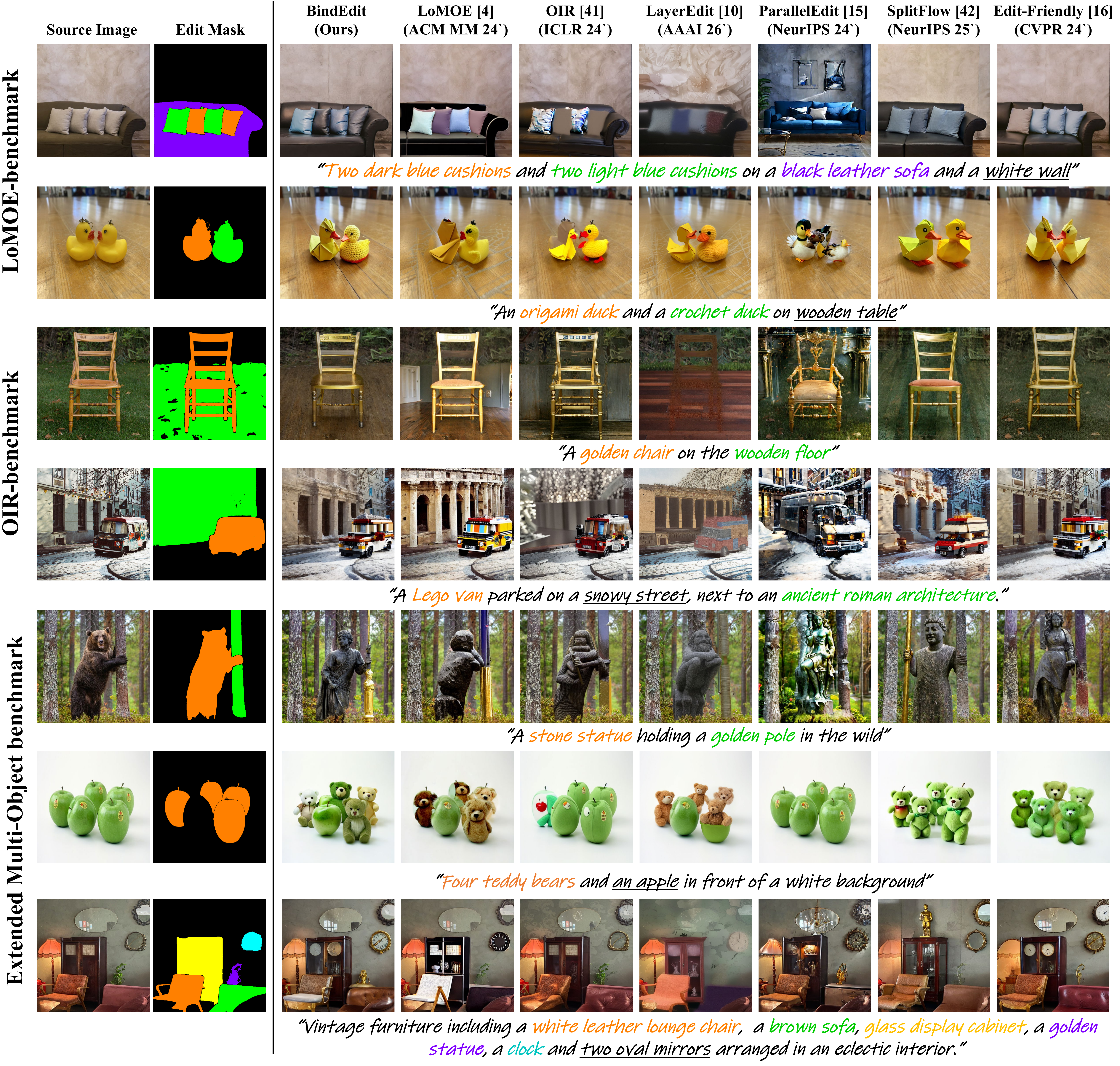}
  \caption{
    \textbf{Qualitative comparisons.}
    We present qualitative results on three benchmarks: LoMOE-Bench, OIR-Bench, and our Extended Multi-Object Benchmark. For each example, the edit masks indicate the regions to be edited, where masks sharing the same color correspond to the same target token group. The target prompt is displayed below each example, with target tokens color-coded to match their corresponding edit masks. Underlined tokens in the target prompt denote source concepts that should be preserved in the edited image.}
  \label{fig:qual}
\end{figure*}

\subsection{Qualitative comparisons}
Fig.~\ref{fig:qual} presents qualitative comparisons with recent multi-object editing methods~\cite{chakrabarty2024lomoe,yang2023object,Fu2025LayerEditDM,Huang2024ParallelEditsEM,Yoon2025SplitFlowFD,huberman2024edit}. We evaluate on existing benchmarks~\cite{chakrabarty2024lomoe,yang2023object} and our proposed multi-object benchmark, which covers a wider range of object counts to assess scalability. Methods that decompose editing into separate per-object operations~\cite{yang2023object,chakrabarty2024lomoe,Fu2025LayerEditDM} can generate intended objects within their respective areas but often leave visible seams or mask residues due to the independent processing of each region. As shown in Row 6, even with the explicit region isolation of LayerEdit~\cite{Fu2025LayerEditDM} or the flow decomposition of SplitFlow~\cite{Yoon2025SplitFlowFD}, artifacts of the source tokens (``an apple'') persist within the edited ``teddy bear'' region. This indicates that spatial or trajectory separation alone is insufficient to suppress attention leakage, often resulting in missing or duplicated edits. In contrast, BindEdit performs the entire edit within a single trajectory.

Furthermore, as shown in Row 1 and Row 7, BindEdit faithfully edits diverse objects ranging from color-specific cushion pairs to complex interior scenes with six or more entities, all without cross-object interference. Even in the densely populated layout of Row 7, BindEdit successfully updates all designated objects while preserving source elements such as ``two oval mirrors'' with high fidelity. These results confirm that BindEdit scales effectively to scenes of increasing complexity while maintaining both coherence and region-level precision.

\subsection{Quantitative results}
\noindent\textbf{Evaluation metrics}
We evaluate all methods using three complementary metrics. \textbf{CLIP-image}\cite{Radford2021LearningTV} measures global text--image alignment between the edited result and the target prompt. \textbf{CLIP-object}\cite{Radford2021LearningTV} computes CLIP similarity between each edited region crop and its corresponding object-level phrase, reflecting how faithfully each individual edit follows its instruction. \textbf{LPIPS}\cite{Zhang2018TheUE} quantifies perceptual similarity between the source and edited images on background regions outside the object masks, measuring how well unedited areas are preserved. 

\begin{table*}[t]
\centering
\caption{\textbf{Quantitative comparisons} on three multi-object editing benchmarks. Best results are in \textbf{bold}, second best are \underline{underlined}. }
\label{tab:quant1}
\resizebox{\linewidth}{!}{
\begin{tabular}{c|c|c|c|c|c|c|c|c}
\hline
\textbf{Metrics}
& \textbf{Datasets} 
& \textbf{BindEdit} 
& \textbf{LoMOE} \cite{chakrabarty2024lomoe}
& \textbf{OIR} \cite{yang2023object}
& \textbf{LayerEdit} \cite{Fu2025LayerEditDM}
& \textbf{ParallelEdits} \cite{Huang2024ParallelEditsEM}
& \textbf{SplitFlow} \cite{Yoon2025SplitFlowFD}
& \textbf{Edit Friendly} \cite{huberman2024edit}
\\ 
\hline
\begin{tabular}{c}
CLIP-obj. $\uparrow$\\
CLIP-img. $\uparrow$\\
LPIPS $\downarrow$
\end{tabular}
& Multi-Object-bench
& \begin{tabular}{c} \textbf{0.2679} \\ \textbf{0.3336} \\ 0.1415 \end{tabular}
& \begin{tabular}{c} 0.2627 \\ 0.3043 \\ \underline{0.0856} \end{tabular}
& \begin{tabular}{c} 0.2523 \\ 0.3028 \\ 0.0947 \end{tabular}
& \begin{tabular}{c} \underline{0.2646} \\ 0.3115 \\ 0.1038 \end{tabular}
& \begin{tabular}{c} 0.2466 \\ 0.2871 \\ \textbf{0.0659} \end{tabular}
& \begin{tabular}{c} 0.2619 \\ \underline{0.3334} \\ 0.2001 \end{tabular}
& \begin{tabular}{c} 0.2595 \\ 0.3274 \\ 0.1737 \end{tabular}
\\
\cline{1-9}
\begin{tabular}{c}
CLIP-obj. $\uparrow$\\
CLIP-img. $\uparrow$\\
LPIPS $\downarrow$
\end{tabular}
& LoMOE-bench \cite{chakrabarty2024lomoe}
& \begin{tabular}{c} \textbf{0.2716} \\ \textbf{0.3341} \\ 0.1604 \end{tabular}
& \begin{tabular}{c} 0.2574 \\ 0.3187 \\ \textbf{0.0903} \end{tabular}
& \begin{tabular}{c} 0.2567 \\ 0.3115 \\ \underline{0.1150} \end{tabular}
& \begin{tabular}{c} \underline{0.2712} \\ 0.3102 \\ 0.1410 \end{tabular}
& \begin{tabular}{c} 0.2483 \\ 0.2831 \\ 0.3162 \end{tabular}
& \begin{tabular}{c} 0.2654 \\ \underline{0.3336} \\ 0.2099 \end{tabular}
& \begin{tabular}{c} 0.2644 \\ 0.3281 \\ 0.1959 \end{tabular}
\\
\cline{1-9}
\begin{tabular}{c}
CLIP-obj. $\uparrow$\\
CLIP-img. $\uparrow$\\
LPIPS $\downarrow$
\end{tabular}
& OIR-bench \cite{yang2023object}
& \begin{tabular}{c} \underline{0.2789} \\ \textbf{0.3448} \\ 0.0923 \end{tabular}
& \begin{tabular}{c} \textbf{0.2808} \\ 0.3368 \\ \textbf{0.0576} \end{tabular}
& \begin{tabular}{c} 0.2716 \\ 0.3339 \\ 0.0723 \end{tabular}
& \begin{tabular}{c} 0.2714 \\ 0.3224 \\ \underline{0.0696} \end{tabular}
& \begin{tabular}{c} 0.2551 \\ 0.3068 \\ 0.1818 \end{tabular}
& \begin{tabular}{c} 0.2746 \\ \underline{0.3416} \\ 0.1224 \end{tabular}
& \begin{tabular}{c} 0.2712 \\ 0.3338 \\ 0.1076 \end{tabular}
\\
\cline{1-9}
\begin{tabular}{c}
CLIP-obj $\uparrow$\\
CLIP-img $\uparrow$\\
LPIPS $\downarrow$
\end{tabular}
& Average
& \begin{tabular}{c} \textbf{0.2728} \\ \textbf{0.3375} \\ 0.1314 \end{tabular}
& \begin{tabular}{c} 0.2670 \\ 0.3199 \\ \textbf{0.0778} \end{tabular}
& \begin{tabular}{c} 0.2602 \\ 0.3161 \\ \underline{0.0940} \end{tabular}
& \begin{tabular}{c} \underline{0.2691} \\ 0.3147 \\ 0.1048 \end{tabular}
& \begin{tabular}{c} 0.2500 \\ 0.2923 \\ 0.1880 \end{tabular}
& \begin{tabular}{c} 0.2673 \\ \underline{0.3362} \\ 0.1775 \end{tabular}
& \begin{tabular}{c} 0.2650 \\ 0.3298 \\ 0.1591 \end{tabular}
\\
\hline
\end{tabular}}
\end{table*}
\begin{table}[t]
\centering
\caption{\textbf{User study results.} Preference rate (\%) across different benchmarks. Best results are in \textbf{bold}.}
\label{tab:user}
\resizebox{\linewidth}{!}{
\begin{tabular}{l|c|c|c|c|c|c}
\hline
\textbf{Dataset}
& \textbf{BindEdit}
& \textbf{LoMOE} \cite{chakrabarty2024lomoe}
& \textbf{OIR} \cite{yang2023object}
& \textbf{ParallelEdits} \cite{Huang2024ParallelEditsEM}
& \textbf{LayerEdit} \cite{Fu2025LayerEditDM}
& \textbf{SplitFlow} \cite{Yoon2025SplitFlowFD}
\\
\hline
LoMOE-bench~\cite{chakrabarty2024lomoe}  & \textbf{67.79\%} & 13.89\% & 3.51\% & 1.65\% & 5.03\% & 8.13\% \\
OIR-bench~\cite{yang2023object}    & \textbf{52.23\%} & 20.47\% & 5.40\% & 3.64\% & 6.03\% & 12.23\% \\
Multi-Object-bench & \textbf{64.44\%} & 15.19\% & 2.10\% & 1.00\% & 12.09\% & 5.18\% \\
\hline
Total        & \textbf{61.47\%} & 16.52\% & 3.54\% & 2.02\% & 8.23\% & 8.22\% \\
\hline
\end{tabular}}
\end{table}

\noindent\textbf{Quantitative comparisons.}
Tab.~\ref{tab:quant1} reports quantitative comparisons across the three benchmarks against six recent methods: LoMOE~\cite{chakrabarty2024lomoe}, OIR~\cite{yang2023object}, LayerEdit~\cite{Fu2025LayerEditDM}, ParallelEdits~\cite{Huang2024ParallelEditsEM}, SplitFlow~\cite{Yoon2025SplitFlowFD}, and Edit-Friendly~\cite{huberman2024edit}.

BindEdit achieves the highest CLIP-image scores on all three datasets, demonstrating its ability to produce globally coherent and visually harmonious scenes. Unlike decompose-and-merge methods that process each object independently, BindEdit edits most objects simultaneously, allowing the denoising process to maintain natural inter-object and background relationships throughout. In addition, BindEdit attains the highest CLIP-object scores on both Multi-Object-bench (0.2679) and LoMOE-bench (0.2716), and ranks second on OIR-bench while maintaining a clear lead in CLIP-image. The consistent CLIP-object improvements are a direct result of our Attention Binding Guidance, which ensures that each target token group is precisely bound to its designated region, preventing the identity blending that degrades per-object alignment in competing methods. On average, BindEdit leads in both CLIP-object (0.2728) and CLIP-image (0.3375), confirming a favorable balance between precise local edits and overall scene quality. Notably, the improvement is most pronounced on Multi-Object-bench, which features scenes with the highest object counts among the three benchmarks. This trend aligns with our motivation. As the number of simultaneously edited objects grows, attention leakage becomes increasingly problematic, and BindEdit's attention-level guidance effectively addresses this challenge. For LPIPS, BindEdit records competitive scores across all benchmarks. LoMOE achieves the lowest LPIPS values, which reflects a conservative editing strategy that keeps outputs closer to the source image, often at the cost of under-editing target objects. BindEdit instead employs soft attention guidance rather than hard mask constraints, allowing edited regions and their surroundings to adapt coherently.

\noindent\textbf{User preference.}
We conduct a user study with 90 participants to evaluate perceptual editing quality, collecting a total of 1,980 responses. Participants were presented with the source image, target prompt, and results from BindEdit and five multi-object image editing methods: OIR~\cite{yang2023object}, LoMOE~\cite{chakrabarty2024lomoe}, LayerEdit~\cite{Fu2025LayerEditDM}, ParallelEdits~\cite{Huang2024ParallelEditsEM}, and SplitFlow~\cite{Yoon2025SplitFlowFD}. For each case, participants were asked to select the image that most accurately reflected the intended modifications while preserving the source image. As reported in Tab.~\ref{tab:user}, BindEdit is preferred in 67.79\%, 52.23\%, and 64.44\% on LoMOE-bench, OIR-bench, and Multi-Object-bench respectively, significantly outperforming the second-best method LoMOE (16.52\% overall). These results confirm that the attention-level constraints in BindEdit lead to edits that are perceived as more natural, prompt-faithful, and coherent by human evaluators.


\begin{figure}[t]
\centering
  \includegraphics[width=1.0\linewidth]{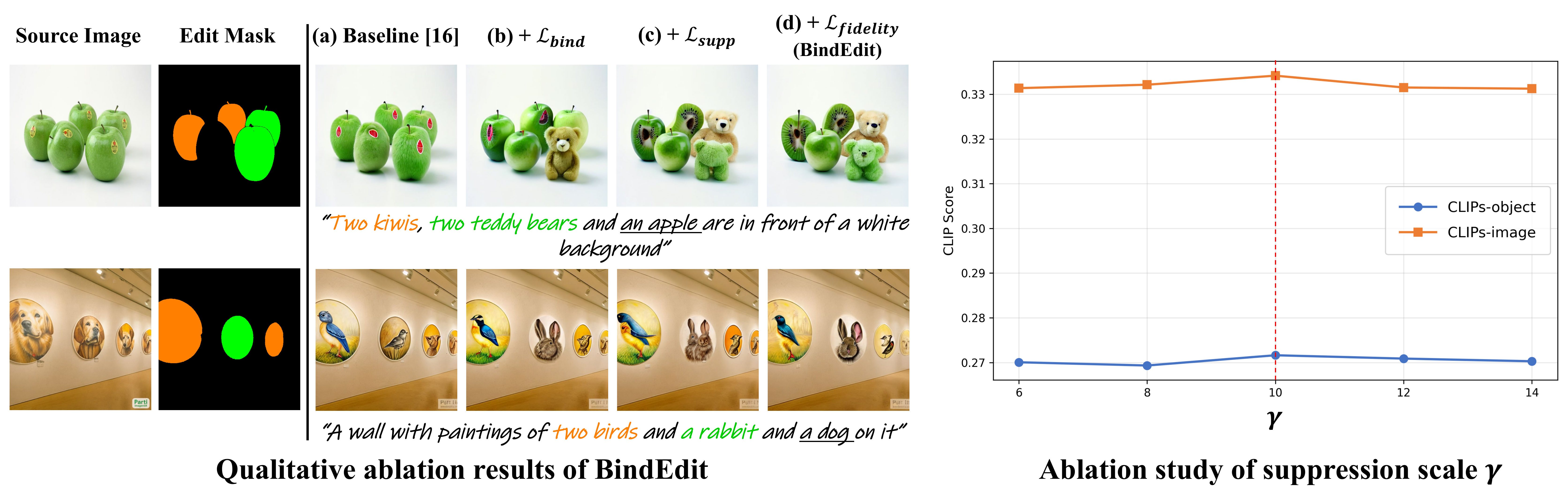}
  \caption{\textbf{Ablation studies.} (Left) Qualitative effect of progressively adding each proposed guidance($\mathcal{L}_{\text{bind}}$,$\mathcal{L}_{\text{supp}}$, and $\mathcal{L}_{\text{fidelity}}$) from left to right. Each loss incrementally resolves object blending, suppresses source token dominance, and consolidates scattered attention into coherent objects. (Right) Robustness to suppression scale $\gamma$ Both CLIP-object and CLIP-image remain stable across $\gamma \in \{ 6,8,10,12,14 \}$.}

  \label{fig:qual_abl}
\end{figure}

\subsection{Ablation Studies}
\label{sec:ablation}

\begin{table}[t]
\centering
\caption{\textbf{Ablation on loss components.} Results on LoMOE-bench. Best in \textbf{bold}, second best \underline{underlined}.}
\label{tab:ablation_loss}
\small
\begin{tabular}{ccc|ccc}
\hline
$\mathcal{L}_{\text{bind}}$ & $\mathcal{L}_{\text{supp}}$ & $\mathcal{L}_{\text{fidelity}}$
& \rotatebox{0}{\scriptsize CLIP-obj.$\uparrow$} 
& \rotatebox{0}{\scriptsize CLIP-img.$\uparrow$} 
& \rotatebox{0}{\scriptsize LPIPS$\downarrow$} \\
\hline
 &  &  & .2663 & .3206 & \textbf{.1499} \\
\checkmark &  &  & .2708 & .3287 & \underline{.1565} \\
\checkmark & \checkmark &  & \textbf{.2718} & \underline{.3321} & .1611 \\
\checkmark & \checkmark & \checkmark & \underline{.2716} & \textbf{.3341} & .1604 \\
\hline
\end{tabular}
\end{table}

We conduct ablation experiments on LoMOE-Bench\cite{chakrabarty2024lomoe} to validate the contribution of each proposed component and the sensitivity to hyperparameter choices.

\noindent\textbf{Effect of each loss component.}
Tab.~\ref{tab:ablation_loss} and Fig.~\ref{fig:qual_abl} (left) present quantitative and qualitative results when progressively adding each loss term. Starting from the baseline~\cite{huberman2024edit}, target token influence fails to localize, causing the intended edits to spread globally across the entire image. Introducing $\mathcal{L}_{\text{bind}}$ successfully grounds the editing intention within the target region, improving both CLIP-object (+1.7\%) and CLIP-image (+2.5\%) over the baseline. However, as visible in Fig.~\ref{fig:qual_abl} (b), traces of the original object (\eg, ``an apple'') still persist within the target mask, indicating residual source dominance. Adding $\mathcal{L}_{\text{supp}}$ effectively suppresses these artifacts, further improving CLIP-object and CLIP-image. The benefit of $\mathcal{L}_{\text{fidelity}}$ is demonstrated in the second example of Fig.~\ref{fig:qual_abl}. Even after suppressing source influence, the target region may exhibit multimodal attention distributions, leading to structural inconsistencies such as a split rabbit artifact in (c). $\mathcal{L}_{\text{fidelity}}$ resolves this by consolidating scattered attention into a single coherent mode while also improving LPIPS. The full BindEdit framework achieves +2.0\% CLIP-object and +4.2\% CLIP-image over the baseline, confirming that each component addresses a distinct aspect of multi-object editing and their combination yields the strongest overall performance.

\noindent\textbf{Ablation study of suppression scale $\gamma$.}
We analyze the sensitivity of the suppression scale $\gamma$ in $\mathcal{L}_{\text{supp}}$, which controls the relative strength of the negative term. As shown in Fig.~\ref{fig:qual_abl} (right), both CLIP-object and CLIP-image remain stable across $\gamma \in \{6, 8, 10, 12, 14\}$, with CLIP-object ranging from 0.2693 to 0.2716 and CLIP-image from 0.3312 to 0.3341. The best performance is achieved at $\gamma = 10$, which we adopt as the default. This narrow variance ($<$1\% across all tested values) confirms that BindEdit is robust to the choice of $\gamma$ within a reasonable range. Additional ablation studies on the loss balancing coefficients ($\lambda_{*}$) are provided in the supplementary material.

\section{Conclusion}

We presented BindEdit, a training-free framework for multi-object image editing that operates within a single diffusion trajectory. Through attention analysis, we identified two key failure modes in multi-object editing: Edit-Token Leakage, where target tokens for different objects become entangled, and Source Dominance Leakage, where residual source tokens suppress newly introduced target concepts. To address these issues, BindEdit introduces three complementary attention-level guidance losses: Attention Binding Guidance jointly regularizes cross- and self-attention for semantic binding and instance isolation, Source Suppression Guidance rebalances cross-attention within editable regions to reduce residual source dominance, and Region Fidelity Regularization consolidates target attention into a coherent spatial distribution. Unlike prior sequential per-object editing methods, BindEdit edits all objects simultaneously within a unified denoising process, achieving higher fidelity and computational efficiency. Experiments on multiple benchmarks, including our proposed multi-object benchmark, demonstrate that BindEdit achieves state-of-the-art performance in automatic metrics and human evaluation while scaling robustly to complex multi-object scenes.
{
    \small
    \bibliographystyle{ieeenat_fullname}
    \bibliography{main}

@String(CVPR= {IEEE Conf. Comput. Vis. Pattern Recog.})

@String(ICCV= {Int. Conf. Comput. Vis.})

@String(ECCV= {Eur. Conf. Comput. Vis.})

@String(TOG= {ACM Trans. Graph.})

@String(AAAI = {AAAI})

@String(CVPR  = {CVPR})

@String(ICCV  = {ICCV})

@String(ECCV  = {ECCV})

@String(TOG   = {ACM TOG})

@article{ho2020denoising,
  title={Denoising diffusion probabilistic models},
  author={Ho, Jonathan and Jain, Ajay and Abbeel, Pieter},
  journal={Advances in neural information processing systems},
  volume={33},
  pages={6840--6851},
  year={2020}
}

@article{song2020denoising,
  title={Denoising diffusion implicit models},
  author={Song, Jiaming and Meng, Chenlin and Ermon, Stefano},
  journal={arXiv preprint arXiv:2010.02502},
  year={2020}
}

@inproceedings{huberman2024edit,
  title={An edit friendly ddpm noise space: Inversion and manipulations},
  author={Huberman-Spiegelglas, Inbar and Kulikov, Vladimir and Michaeli, Tomer},
  booktitle={Proceedings of the IEEE/CVF Conference on Computer Vision and Pattern Recognition},
  pages={12469--12478},
  year={2024}
}

@article{hertz2022prompt,
  title={Prompt-to-prompt image editing with cross attention control},
  author={Hertz, Amir and Mokady, Ron and Tenenbaum, Jay and Aberman, Kfir and Pritch, Yael and Cohen-Or, Daniel},
  journal={arXiv preprint arXiv:2208.01626},
  year={2022}
}

@inproceedings{tumanyan2023plug,
  title={Plug-and-play diffusion features for text-driven image-to-image translation},
  author={Tumanyan, Narek and Geyer, Michal and Bagon, Shai and Dekel, Tali},
  booktitle={Proceedings of the IEEE/CVF conference on computer vision and pattern recognition},
  pages={1921--1930},
  year={2023}
}

@inproceedings{mokady2023null,
  title={Null-text inversion for editing real images using guided diffusion models},
  author={Mokady, Ron and Hertz, Amir and Aberman, Kfir and Pritch, Yael and Cohen-Or, Daniel},
  booktitle={Proceedings of the IEEE/CVF conference on computer vision and pattern recognition},
  pages={6038--6047},
  year={2023}
}

@inproceedings{hertz2023delta,
  title={Delta denoising score},
  author={Hertz, Amir and Aberman, Kfir and Cohen-Or, Daniel},
  booktitle={Proceedings of the IEEE/CVF International Conference on Computer Vision},
  pages={2328--2337},
  year={2023}
}

@inproceedings{ren2025fds,
  title={FDS: Frequency-Aware Denoising Score for Text-Guided Latent Diffusion Image Editing},
  author={Ren, Yufan and Jiang, Zicong and Zhang, Tong and Forchhammer, S{\o}ren and S{\"u}sstrunk, Sabine},
  booktitle={Proceedings of the Computer Vision and Pattern Recognition Conference},
  pages={2651--2660},
  year={2025}
}

@inproceedings{chakrabarty2024lomoe,
  title={Lomoe: Localized multi-object editing via multi-diffusion},
  author={Chakrabarty, Goirik and Chandrasekar, Aditya and Hebbalaguppe, Ramya and AP, Prathosh},
  booktitle={Proceedings of the 32nd ACM International Conference on Multimedia},
  pages={3342--3351},
  year={2024}
}

@article{yang2023object,
  title={Object-aware inversion and reassembly for image editing},
  author={Yang, Zhen and Ding, Ganggui and Wang, Wen and Chen, Hao and Zhuang, Bohan and Shen, Chunhua},
  journal={arXiv preprint arXiv:2310.12149},
  year={2023}
}

@article{Saharia2022PhotorealisticTD,
  title={Photorealistic Text-to-Image Diffusion Models with Deep Language Understanding},
  author={Chitwan Saharia and William Chan and Saurabh Saxena and Lala Li and Jay Whang and Emily L. Denton and Seyed Kamyar Seyed Ghasemipour and Burcu Karagol Ayan and Seyedeh Sara Mahdavi and Raphael Gontijo Lopes and Tim Salimans and Jonathan Ho and David J. Fleet and Mohammad Norouzi},
  journal={ArXiv},
  year={2022},
  volume={abs/2205.11487},
  url={https://api.semanticscholar.org/CorpusID:248986576}
}

@article{Peebles2022ScalableDM,
  title={Scalable Diffusion Models with Transformers},
  author={William S. Peebles and Saining Xie},
  journal={2023 IEEE/CVF International Conference on Computer Vision (ICCV)},
  year={2022},
  pages={4172-4182},
  url={https://api.semanticscholar.org/CorpusID:254854389}
}

@article{Labs2025FLUX1KF,
  title={FLUX.1 Kontext: Flow Matching for In-Context Image Generation and Editing in Latent Space},
  author={Black Forest Labs and Stephen Batifol and A. Blattmann and Frederic Boesel and Saksham Consul and Cyril Diagne and Tim Dockhorn and Jack English and Zion English and Patrick Esser and Sumith Kulal and Kyle Lacey and Yam Levi and Cheng Li and Dominik Lorenz and Jonas Muller and Dustin Podell and Robin Rombach and Harry Saini and Axel Sauer and Luke Smith},
  journal={ArXiv},
  year={2025},
  volume={abs/2506.15742},
  url={https://api.semanticscholar.org/CorpusID:279464475}
}

@article{Rombach2021HighResolutionIS,
  title={High-Resolution Image Synthesis with Latent Diffusion Models},
  author={Robin Rombach and A. Blattmann and Dominik Lorenz and Patrick Esser and Bj{\"o}rn Ommer},
  journal={2022 IEEE/CVF Conference on Computer Vision and Pattern Recognition (CVPR)},
  year={2021},
  pages={10674-10685},
  url={https://api.semanticscholar.org/CorpusID:245335280}
}

@article{Podell2023SDXLIL,
  title={SDXL: Improving Latent Diffusion Models for High-Resolution Image Synthesis},
  author={Dustin Podell and Zion English and Kyle Lacey and A. Blattmann and Tim Dockhorn and Jonas Muller and Joe Penna and Robin Rombach},
  journal={ArXiv},
  year={2023},
  volume={abs/2307.01952},
  url={https://api.semanticscholar.org/CorpusID:259341735}
}

@article{Zhu2025MDEEditMD,
  title={MDE-Edit: Masked Dual-Editing for Multi-Object Image Editing via Diffusion Models},
  author={Hongyang Zhu and Haipeng Liu and Bo Fu and Yang Wang},
  journal={ArXiv},
  year={2025},
  volume={abs/2505.05101},
  url={https://api.semanticscholar.org/CorpusID:278394381}
}

@article{Kang2025BoundedEM,
  title={Bounded Editing: Multi-Object Image Manipulation with Region-Specific Control},
  author={Min Gyu Kang and Keon Kim and Yong Suk Choi},
  journal={Proceedings of the 40th ACM/SIGAPP Symposium on Applied Computing},
  year={2025},
  url={https://api.semanticscholar.org/CorpusID:278603399}
}

@article{Chen2023TrainingFreeLC,
  title={Training-Free Layout Control with Cross-Attention Guidance},
  author={Minghao Chen and Iro Laina and Andrea Vedaldi},
  journal={2024 IEEE/CVF Winter Conference on Applications of Computer Vision (WACV)},
  year={2023},
  pages={5331-5341},
  url={https://api.semanticscholar.org/CorpusID:258041377}
}

@article{Dahary2024BeYB,
  title={Be Yourself: Bounded Attention for Multi-Subject Text-to-Image Generation},
  author={Omer Dahary and Or Patashnik and Kfir Aberman and Daniel Cohen-Or},
  journal={In European Conference on Computer Vision (ECCV)},
  year={2024}
}

@article{Han2025SpatialTO,
  title={Spatial Transport Optimization by Repositioning Attention Map for Training-Free Text-to-Image Synthesis},
  author={Woojung Han and Yeonkyung Lee and Chanyoung Kim and Kwanghyun Park and Seong Jae Hwang},
  journal={2025 IEEE/CVF Conference on Computer Vision and Pattern Recognition (CVPR)},
  year={2025},
  pages={18401-18410},
  url={https://api.semanticscholar.org/CorpusID:277435518}
}

@article{Qiu2024SelfCrossDG,
  title={Self-Cross Diffusion Guidance for Text-to-Image Synthesis of Similar Subjects},
  author={Weimin Qiu and Jieke Wang and Meng Tang},
  journal={2025 IEEE/CVF Conference on Computer Vision and Pattern Recognition (CVPR)},
  year={2024},
  pages={23528-23538},
  url={https://api.semanticscholar.org/CorpusID:274422699}
}

@article{Yoon2025SplitFlowFD,
  title={SplitFlow: Flow Decomposition for Inversion-Free Text-to-Image Editing},
  author={Sunghoon Yoon and Minghan Li and Gaspard Beaudouin and Congcong Wen and Muhammad Rafay Azhar and Mengyu Wang},
  journal={ In Conference on Neural
Information Processing Systems (NeurIPS)},
  year={2025}
}

@article{Fu2025LayerEditDM,
  title={LayerEdit: Disentangled Multi-Object Editing via Conflict-Aware Multi-Layer Learning},
  author={Fengyi Fu and Mengqi Huang and Lei Zhang and Zhendong Mao},
  journal={The 40th Annual AAAI Conference on Artificial Intelligence},
  year={2026},
}

@article{Huang2024ParallelEditsEM,
  title={ParallelEdits: Efficient Multi-Aspect Text-Driven Image Editing with Attention Grouping},
  author={Mingzhen Huang and Jialing Cai and Shan Jia and Vishnu Suresh Lokhande and Siwei Lyu},
  journal={Advances in Neural Information Processing Systems 37},
  year={2024},
  url={https://api.semanticscholar.org/CorpusID:270216857}
}

@article{Brack2023LEDITSLI,
  title={LEDITS++: Limitless Image Editing Using Text-to-Image Models},
  author={Manuel Brack and Felix Friedrich and Katharina Kornmeier and Linoy Tsaban and Patrick Schramowski and Kristian Kersting and Apolin'ario Passos},
  journal={2024 IEEE/CVF Conference on Computer Vision and Pattern Recognition (CVPR)},
  year={2023},
  pages={8861-8870},
  url={https://api.semanticscholar.org/CorpusID:265466786}
}

@article{Li2025MoEditOL,
  title={MoEdit: On Learning Quantity Perception for Multi-object Image Editing},
  author={Yanfeng Li and Ka-Hou Chan and Yue Sun and Chan Quan Loi Lam and Tong Tong and Zitong Yu and Keren Fu and Xiaohong Liu and Tao Tan},
  journal={2025 IEEE/CVF Conference on Computer Vision and Pattern Recognition (CVPR)},
  year={2025},
  pages={2683-2693},
  url={https://api.semanticscholar.org/CorpusID:276961755}
}

@article{Couairon2022DiffEditDS,
  title={DiffEdit: Diffusion-based semantic image editing with mask guidance},
  author={Guillaume Couairon and Jakob Verbeek and Holger Schwenk and Matthieu Cord},
  journal={ArXiv},
  year={2022},
  volume={abs/2210.11427},
  url={https://api.semanticscholar.org/CorpusID:253018768}
}

@inproceedings{Meng2021SDEditGI,
  title={SDEdit: Guided Image Synthesis and Editing with Stochastic Differential Equations},
  author={Chenlin Meng and Yutong He and Yang Song and Jiaming Song and Jiajun Wu and Jun-Yan Zhu and Stefano Ermon},
  booktitle={International Conference on Learning Representations},
  year={2021},
  url={https://api.semanticscholar.org/CorpusID:245704504}
}

@article{Cao2023MasaCtrlTM,
  title={MasaCtrl: Tuning-Free Mutual Self-Attention Control for Consistent Image Synthesis and Editing},
  author={Ming Cao and Xintao Wang and Zhongang Qi and Ying Shan and Xiaohu Qie and Yinqiang Zheng},
  journal={2023 IEEE/CVF International Conference on Computer Vision (ICCV)},
  year={2023},
  pages={22503-22513},
  url={https://api.semanticscholar.org/CorpusID:258179432}
}

@article{Wu2024FreeDiffPF,
  title={FreeDiff: Progressive Frequency Truncation for Image Editing with Diffusion Models},
  author={Wei Wu and Qingnan Fan and Shuai Qin and Hong Gu and Ruoyu Zhao and Antoni B. Chan},
  journal={In European Conference on Computer Vision (ECCV)},
  year={2024},
}

@article{Koo2024FlexiEditFL,
  title={FlexiEdit: Frequency-Aware Latent Refinement for Enhanced Non-Rigid Editing},
  author={Gwanhyeong Koo and Sunjae Yoon and Jiajing Hong and Changdong Yoo},
  journal={In European Conference on Computer Vision (ECCV)},
  year={2024}
}

@article{Liu2024TowardsUC,
  title={Towards Understanding Cross and Self-Attention in Stable Diffusion for Text-Guided Image Editing},
  author={Bingyan Liu and Chengyu Wang and Tingfeng Cao and Kui Jia and Jun Huang},
  journal={2024 IEEE/CVF Conference on Computer Vision and Pattern Recognition (CVPR)},
  year={2024},
  pages={7817-7826},
  url={https://api.semanticscholar.org/CorpusID:268253430}
}

@article{Nam2023ContrastiveDS,
  title={Contrastive Denoising Score for Text-Guided Latent Diffusion Image Editing},
  author={Hyelin Nam and Gihyun Kwon and Geon Yeong Park and Jong Chul Ye},
  journal={2024 IEEE/CVF Conference on Computer Vision and Pattern Recognition (CVPR)},
  year={2023},
  pages={9192-9201},
  url={https://api.semanticscholar.org/CorpusID:265506264}
}

@article{Chefer2023AttendandExciteAS,
  title={Attend-and-Excite: Attention-Based Semantic Guidance for Text-to-Image Diffusion Models},
  author={Hila Chefer and Yuval Alaluf and Yael Vinker and Lior Wolf and Daniel Cohen-Or},
  journal={ACM Transactions on Graphics (TOG)},
  year={2023},
  volume={42},
  pages={1 - 10},
  url={https://api.semanticscholar.org/CorpusID:256416326}
}

@article{Tian2023DiffuseAA,
  title={Diffuse, Attend, and Segment: Unsupervised Zero-Shot Segmentation using Stable Diffusion},
  author={Junjiao Tian and Lavisha Aggarwal and Andrea Colaco and Zsolt Kira and Mar Gonz{\'a}lez-Franco},
  journal={2024 IEEE/CVF Conference on Computer Vision and Pattern Recognition (CVPR)},
  year={2023},
  pages={3554-3563},
  url={https://api.semanticscholar.org/CorpusID:261101006}
}

@article{Avrahami2024StableFV,
  title={Stable Flow: Vital Layers for Training-Free Image Editing},
  author={Omri Avrahami and Or Patashnik and Ohad Fried and Egor Nemchinov and Kfir Aberman and Dani Lischinski and Daniel Cohen-Or},
  journal={2025 IEEE/CVF Conference on Computer Vision and Pattern Recognition (CVPR)},
  year={2024},
  pages={7877-7888},
  url={https://api.semanticscholar.org/CorpusID:274165561}
}

@article{Kulikov2024FlowEditIT,
  title={FlowEdit: Inversion-Free Text-Based Editing Using Pre-Trained Flow Models},
  author={Vladimir Kulikov and Matan Kleiner and Inbar Huberman-Spiegelglas and Tomer Michaeli},
  journal={ArXiv},
  year={2024},
  volume={abs/2412.08629},
  url={https://api.semanticscholar.org/CorpusID:274638560}
}

@article{Shin2025ExploringMD,
  title={Exploring Multimodal Diffusion Transformers for Enhanced Prompt-based Image Editing},
  author={Joonghyuk Shin and Alchan Hwang and Yujin Kim and Daneul Kim and Jaesik Park},
  journal={ArXiv},
  year={2025},
  volume={abs/2508.07519},
  url={https://api.semanticscholar.org/CorpusID:280566537}
}

@article{Chen2024RegionAwareTG,
  title={Region-Aware Text-to-Image Generation via Hard Binding and Soft Refinement},
  author={Zhennan Chen and Yajie Li and Haofan Wang and Zhibo Chen and Zhengkai Jiang and Jun Li and Qian Wang and Jian Yang and Ying Tai},
  journal={ArXiv},
  year={2024},
  volume={abs/2411.06558},
  url={https://api.semanticscholar.org/CorpusID:273962829}
}

@inproceedings{Park2025CrossAttentionHP,
  title={Cross-Attention Head Position Patterns Can Align with Human Visual Concepts in Text-to-Image Generative Models},
  author={Jungwon Park and Jungmin Ko and Dongnam Byun and Jangwon Suh and Wonjong Rhee},
  booktitle={International Conference on Learning Representations},
  year={2025},
  url={https://api.semanticscholar.org/CorpusID:279993085}
}

@article{Li2023DivideB,
  title={Divide \& Bind Your Attention for Improved Generative Semantic Nursing},
  author={Yumeng Li and Margret Keuper and Dan Zhang and Anna Khoreva},
  journal={ArXiv},
  year={2023},
  volume={abs/2307.10864},
  url={https://api.semanticscholar.org/CorpusID:259991537}
}

@inproceedings{Tang2022WhatTD,
  title={What the DAAM: Interpreting Stable Diffusion Using Cross Attention},
  author={Raphael Tang and Akshat Pandey and Zhiying Jiang and Gefei Yang and Karun Kumar and Jimmy J. Lin and Ferhan Ture},
  booktitle={Annual Meeting of the Association for Computational Linguistics},
  year={2022},
  url={https://api.semanticscholar.org/CorpusID:252780146}
}

@inproceedings{Radford2021LearningTV,
  title={Learning Transferable Visual Models From Natural Language Supervision},
  author={Alec Radford and Jong Wook Kim and Chris Hallacy and Aditya Ramesh and Gabriel Goh and Sandhini Agarwal and Girish Sastry and Amanda Askell and Pamela Mishkin and Jack Clark and Gretchen Krueger and Ilya Sutskever},
  booktitle={International Conference on Machine Learning},
  year={2021},
  url={https://api.semanticscholar.org/CorpusID:231591445}
}

@article{Zhang2018TheUE,
  title={The Unreasonable Effectiveness of Deep Features as a Perceptual Metric},
  author={Richard Zhang and Phillip Isola and Alexei A. Efros and Eli Shechtman and Oliver Wang},
  journal={2018 IEEE/CVF Conference on Computer Vision and Pattern Recognition},
  year={2018},
  pages={586-595},
  url={https://api.semanticscholar.org/CorpusID:4766599}
}
}


\end{document}